%% file: main.tex
\documentclass[lettersize,journal]{IEEEtran}

\usepackage{fancyhdr, epsfig, amsthm, amsmath, amssymb, amsfonts, subfigure,dsfont,float}

\usepackage[utf8]{inputenc} 
\usepackage[T1]{fontenc}
\usepackage{url}  

\usepackage{cite}
\usepackage{algorithmic}
\usepackage{graphicx}
\usepackage{textcomp}
\usepackage[table,xcdraw]{xcolor}
\usepackage{tabularx}
\usepackage{tikz}
\usepackage{enumitem}
\usepackage{url}
\usepackage{booktabs}
\usepackage{array}
\usepackage{multirow}
\usepackage{booktabs} 
\usepackage{siunitx} 
\usepackage{xfrac}
\usepackage{nicefrac}
\usepackage{psfrag}
\usepackage{times}
\usepackage{balance} 
\usepackage{bm}
\usepackage[bottom]{footmisc}
\usepackage{mwe}
\usepackage{color}
\usepackage{booktabs,multirow}
\usetikzlibrary{patterns,arrows}

\usepackage[a4paper, top=0.6in, bottom=0.7in, left=0.55in, right=0.55in]{geometry}

\input{abbreviations}
\input{my_definations}

\begin{document}

\title{Semantic Communication and Control Co-Design for Multi-Objective Distinct Systems}

\author{
Abanoub M. Girgis,~\IEEEmembership{Student Member,~IEEE},
Hyowoon~Seo,~\IEEEmembership{Member,~IEEE},
and Mehdi~Bennis,~\IEEEmembership{Fellow,~IEEE}
\thanks{This research was supported in part by the Research Council of Finland (former Academy of Finland) $6$G Flagship Programme (Grant Number: 346208) and project SMARTER, projects EU-ICT IntellIoT and EU-CHISTERA LeadingEdge, and CENTRIC (Grant Agreement 101096379), and VERGE (Grant Agreement 101096034).}
\thanks{A. M. Girgis and M. Bennis are with the Centre for Wireless Communications, University of Oulu, Oulu 90014, Finland (e-mail: abanoub.pipaoy@oulu.fi; mehdi.bennis@oulu.fi).}
\thanks{H. Seo is with the Department of Electronics and Communications Engineering, Kwangwoon University, Seoul, South Korea (e-mail: hyowoonseo@kw.ac.kr).}
}

\maketitle 

\begin{abstract}
This letter introduces a deep learning approach to predict the semantic control dynamics of different systems with different dynamics and control rules. 
By leveraging the Koopman operator in an autoencoder (AE) framework, the system's state evolution is linearized in the latent space using a dynamic semantic Koopman (DSK) model, capturing the baseline semantic dynamics. 
The signal temporal logic (STL) is incorporated through a logical semantic Koopman (LSK) model to encode system-specific control rules. 
These models form the proposed semantic logical Koopman AE that reduces communication and computational costs without compromising control performance, showing a $91.65\%$ reduction in communication samples and significant performance gains in simulation. 
\end{abstract}


\input{Sec1_Introduction.tex}
\input{Sec2_System_Model.tex}

\input{Sec3_Prelimiaries.tex}
\input{Sec4_CLD_Koopman_AE}

\input{Sec5_Simulation_Results.tex}

\input{Sec6_Conclusion.tex}

\bibliographystyle{IEEEtran}
\bibliography{IEEEabrv,bibliography}

\end{document}

%% file: abbreviations.tex
\usepackage[acronym]{glossaries}

\newacronym{wncs}{WNCS}{wireless networked control systems}

\newacronym{iot}{IoT}{internet-of-things}

\newacronym{iiot}{IIoT}{industrial internet-of-things}

\newacronym{cococo}{CoCoCo}{communication and control co-design}

\newacronym{ml}{ML}{machine learning}

\newacronym{dml}{DML}{distributed machine learning}

\newacronym{qos}{QoS}{quality-of-service}

\newacronym{qoe}{QoE}{quality-of-experience}

\newacronym{snr}{SNR}{signal-to-noise ratio}

\newacronym{urllc}{URLLC}{ultra-reliable low-latency communication}

\newacronym{gpr}{GPR}{Gaussian process regression}

\newacronym{lqr}{LQR}{linear quadratic regulator}

\newacronym{mse}{MSE}{mean square error}

\newacronym{mmse}{MMSE}{minimum mean square error}

\newacronym{aoi}{AoI}{age of information}

\newacronym{gan}{GAN}{generative adversarial network}

\newacronym{dnn}{DNN}{deep neural network}
 
\newacronym{rl}{RL}{Reinforcement learning}

\newacronym{dl}{DL}{deep learning}

\newacronym{stl}{STL}{signal temporal logic}
 
\newacronym{iid}{i.i.d.}{independent and identically distributed} 
 
\newacronym{ae}{AE}{auto-encoder}

\newacronym{tdma}{TDMA}{time-division multiple access}
      
\newacronym{nn}{NN}{neural network}

\newacronym{rnns}{RNNs}{recurrent neural networks}

\newacronym{lstm}{LSTM}{long short-term memory}
  
\newacronym{relu}{ReLu}{rectified linear unit} 
  
\newacronym{nrmse}{NRMSE}{normalized root mean square error}

\newacronym{cdf}{cdf}{cumulative distribution function}
 
\newacronym{uav}{UAV}{unmanned aerial vehicle}
   
\newacronym{mld}{MLD}{mixed logical dynamical}

\newacronym{mimo}{MIMO}{Multi-Input Multi-Output}

\newacronym{tdd}{TDD}{time-division duplex}

\newacronym{lsk}{LSK}{logic semantics Koopman}

\newacronym{dsk}{DSK}{dynamic semantics Koopman}

\newacronym{6g}{6G}{sixth generation}

\newacronym{sc}{SC}{semantic communication}

\newacronym{goc}{GoC}{goal-oriented communication}

%% file: my_definations.tex
\definecolor{color1}{RGB}{237, 191, 193}
\definecolor{color2}{RGB}{229, 153, 157}
\definecolor{color3}{RGB}{225, 123, 116}
\definecolor{color4}{RGB}{217,87,77}
\definecolor{color5}{RGB}{203,52,38}

\def\l{\left}
\def\r{\right}
\def\({\l(}
\def\){\r)}
\def\[{\l[}
\def\]{\r]}

\def\BibTeX{{\rm B\kern-.05em{\sc i\kern-.025em b}\kern-.08em
    T\kern-.1667em\lower.7ex\hbox{E}\kern-.125emX}}

%% file: Sec1_Introduction.tex
\section{Introduction}
\label{Sec1}

\IEEEPARstart{T}{he} rapid advancement of communication and control technologies underpins the development of emerging \gls{6g} applications, such as autonomous systems and industrial automation~\cite{ zeng2019joint, getu2024survey}. 
However, traditional system architectures treat communication and control separately, leading to scalability bottlenecks and stability challenges in distributed and resource-constrained environments. 
To bridge this gap, \gls{cococo} has emerged as a promising paradigm that optimizes communication resources without compromising control performance. 
Existing \gls{cococo} approaches primarily focus on adaptive scheduling and resource allocation based on channel and control states~\cite{girgis2021predictive, eisen2022communication, zhao2018toward}, yet they often overlook the underlying explicit system dynamics that govern multi-objective control systems and rely on lossy compression techniques to reduce communication overhead.

In parallel, the emergence of \gls{goc} and \gls{sc} has reshaped perspectives on information transmission for \gls{6g} networks. 
\gls{goc} prioritizes the transmission of task-relevant information required to achieve specific control or decision-making objectives, while \gls{sc} focuses on extracting and transmitting meaningful representations from high-dimensional data that retain essential semantics of the information~\cite{seo2021semantics, strinati2024goal, xie2021deep, 10849550}.  
Leveraging deep architectures such as \gls{ae}, \gls{sc} enables the extraction and transmission of low-dimensional semantic representations, sufficient to accomplish tasks at the receiver.  
By transmitting only what is necessary to accomplish tasks, \gls{goc} directly enhances the effectiveness of achieving task-specific control objectives, while \gls{sc} improves communication efficiency by eliminating redundant information. 
Building upon these, \emph{semantic \gls{cococo}} emerges as an integrated concept that simultaneously benefits from both communication approaches: aligns with \gls{goc} to maximize task-specific control effectiveness and aligns with \gls{sc} to minimize communication overhead.  
This synergy makes semantic \gls{cococo} particularly suitable for large-scale multi-objective distributed systems operating under limited communication and computational resources. 
Nonetheless, key challenges remain in jointly optimizing communication cost, control performance, and scalability across heterogeneous systems~\cite{girgis2023semantic}. 

The multi-objective distinct control systems have been investigated in the literature. 
Existing approaches often employ control-aware scheduling and resource allocation to determine when and which control systems are granted transmission opportunities~\cite{eisen2019control, lima2020learning, de2021age}. 
Specifically, these approaches prioritize control performance by dynamically allocating wireless resources based on control states or channel conditions, ensuring that the most critical control systems maintain acceptable performance under limited wireless resources. 
However, such approaches typically treat each control system separately rather than capturing cross-system interdependencies. 
More recently, multi-objective reinforcement learning has been explored to coordinate multiple distinct control tasks with shared wireless resources~\cite{liu2014multiobjective,hayes2021practical}.  
It enables agents to learn policies that balance different objectives through reward shaping.
Although it adapts to changing environments and jointly optimizes multiple objectives, it requires extensive interaction samples that lead to high communication and computational overhead, and it struggles to capture the underlying dynamics that couple among heterogeneous control systems.

Our prior work in~\cite{girgis2023semantic} introduced a semantic \gls{cococo} framework for multi-objective control systems sharing the same dynamics but differing control rules, demonstrating that a correlated semantic structure significantly reduces communication overhead. 
To this end, we extend this approach to a more complex and realistic setting: \emph{multi-objective control of systems with both distinct dynamics and control rules.}
Specifically, we propose a novel semantic \gls{cococo} framework based on Koopman operator theory~\cite{brunton2016koopman} for globally linearizing non-linear dynamics within latent space. 
Our proposed framework employs a \gls{dsk} model to extract a shared semantic structure from a reference system and a \gls{lsk} model to encode system-specific control rules using \gls{stl}~\cite{leung2019backpropagation}. 
The \gls{ae} is pre-trained on the reference system and then fine-tuned using a small subset of data from each system to adapt to local dynamic variations. 
The composition of \gls{dsk} and \gls{lsk} forms the proposed \emph{semantic logical Koopman \gls{ae}}, a scalable, task-driven approach that preserves control performance while reducing communication costs.

Unlike~\cite{girgis2023semantic}, which is based on systems with identical system dynamics, the proposed approach addresses the scalability challenges in heterogeneous systems, enabling knowledge transfer across structurally distinct systems governed by logical control rules.
Each system's \gls{lsk} model is trained using a temporal logic-based loss, allowing control rules to be embedded directly into latent dynamics. 
This design eliminates the need to train a full Koopman model per system, reducing computational and communication overhead. 
Extensive simulations on five distinct cart-pole systems validate the proposed approach, showing a $75.78\%$ improvement in average state prediction, a $91.65\%$ reduction in communication samples, and a $92.91 \%$ increase in control performance under $15 \, \mathrm{dB}$ \gls{snr} and using two-dimensional state representations. 

%% file: Sec2_System_Model.tex
\section{Preliminaries}
\label{Sec2}

\subsection{System Model}
\begin{figure}[t]
   \centering
   \includegraphics[trim={1cm 1.5cm 1cm 1.0cm}, clip, width=8.75cm, height=5.3cm]{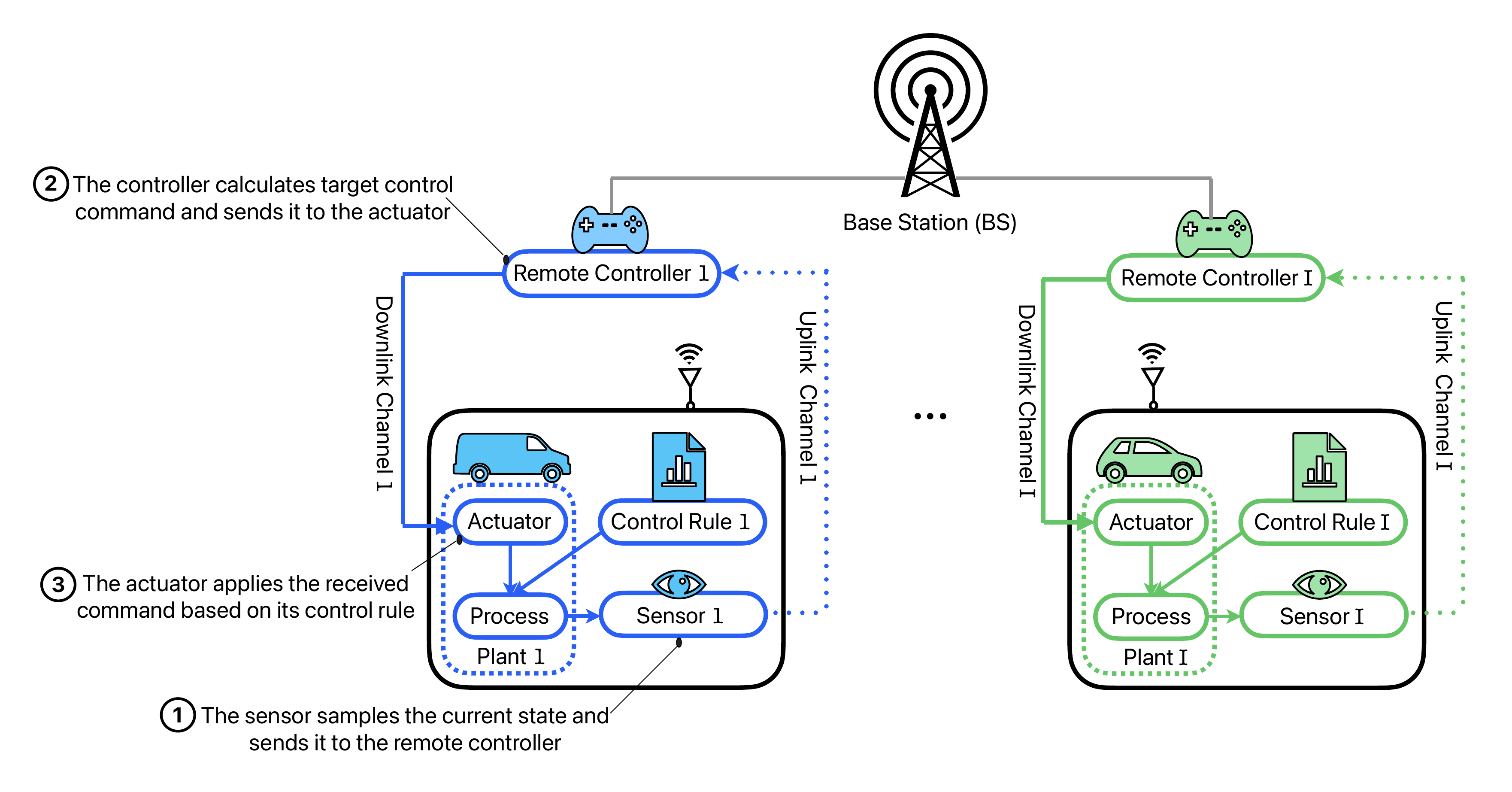} \\
   \caption{Illustration of multi-objective distinct and distributed control systems.} \label{fig_MCS}
\end{figure}

Consider controlling multi-objective distinct control systems composed of plants, sensors, and remote controllers, as depicted in Fig.~\ref{fig_MCS}. 
The set of plants is denoted by $\mathcal{I}$, where each plant represents a non-linear dynamic \emph{process} controlled by an \emph{actuator}, with sensors sampling the process state.
These sampled states are sent to a remote controller with high computational power to calculate the target control commands.  
The commands are then transmitted back to the actuators to drive the process to the desired behavior. 
The sensors and actuators are co-located and share a transceiver for wireless communication, while the remote controllers communicate with the sensors via wireless channels.  
Each plant follows different \textit{control rules} to meet objectives such as stability and safety. 
Further details on the control and communication systems are provided next.

\subsubsection{Control Model}
\label{Sec2_1}

Each plant sensor samples its $p$-dimensional state at a sampling rate $\tau_o$ and transmits these to the remote controller over a wireless channel.
For plant $i \in \mathcal{I}$, the state at time $t = k \tau_o$ is given by $\mathbf{x}_{i,k} \in \mathbb{R}^{p}$.
Upon receiving $\mathbf{x}_{i,k}$, the remote controller computes an optimal control command $\mathbf{u}_{i,k} \in \mathbb{R}^{q}$ and sends it to the actuator, which then influences the plant state. 
The process state evolves as a control-affine non-linear dynamics as~\cite{kaiser2021data} 
\begin{align}
    \label{eq_control_affine}
    \mathbf{x}_{i,k+1} =  \mathbf{f}^{s}_{\varphi_i}( \mathbf{x}_{i,k} ) + \mathbf{F}^{u}_{\varphi_i} ( \mathbf{x}_{i,k} ) \mathbf{u}_{i,k}  + \mathbf{n}_{s,k},
\end{align}
under given control rules $\varphi_i$, where $\mathbf{n}_{s,k} \in \mathbb{R}^{p}$ is system noise, modeled as \gls{iid} Gaussian random vector.
$\mathbf{f}^{s}_{\varphi_i}$ represents the non-linear state transition function and $\mathbf{F}^{u}_{\varphi_i}$ is the state-dependent control function, both are subject to predefined control rules $\varphi_{i}$.   
Each control rule, $\varphi_i$, will be formalized as an \gls{stl} formula, translating natural language rules into a mathematical format.
This concept and its integration into the semantic logical Koopman \gls{ae} are discussed further in Section~\ref{Sec4}.

\subsubsection{Wireless Communication}
\label{Sec2_2}

Each plant communicates with the remote controller using a dedicated orthogonal channel.
Both uplink (sensor-to-controller) and downlink (controller-to-actuator) transmissions share the same channel, alternating via a time-division duplex. 
The channel reciprocity ensures consistent transmission power $P_i$ in both directions. 
Wireless channels are modeled using path loss and Rayleigh fading, with path loss defined as \vspace{-5pt}
\begin{align}
    \label{eq_path_loss}
     \mathrm{PL}_{\mathrm{dB}} \left( D \right)  = \mathrm{PL}_{\mathrm{dB}} \left( D_0 \right)  + 10 \eta\log_{10} \left( \frac{D}{D_0} \right),
\end{align}
where $D$ is the transmitter-receiver distance, $\mathrm{PL}_{\mathrm{dB}} \left(D_0\right)$ is the path loss at the reference distance $D_{0}$, and $\eta \geq 2$ is the path loss exponent. 
The received \gls{snr} for plant $i$ at time $k$ is expressed as 
\begin{align}
     \label{eq_SNR}
     \gamma_{i,k} = 10^{-\frac{\mathrm{PL}_{\mathrm{dB}}\left( D_o \right)}{10}}\frac{P_i | H_{i,k}|^{2}}{ N_c }\left(\frac{D_o}{D}\right)^{\eta},
\end{align}
where $H_{i,k}$ is the Rayleigh fading channel, and $N_c$ is the power of additive white Gaussian noise. 
The channel capacity for plant $i$ at time $k$ is expressed as 
\begin{align}
     \label{eq_Shannon_data_rate}
     R_{i,k} = B \log_{2} \left( 1 + \gamma_{i,k} \right),
\end{align} 
with the outage probability given by 
\begin{align}
{  \epsilon_{i,k} } = 1 - \text{exp} \left[ -10^{\frac{\mathrm{PL}_{\mathrm{dB}}(D_0)}{10}} \frac{ N_c D^{\eta} }{P_i} \left(2^{ \frac{\bar{R}}{W } } -1\right)  \right]. \label{eq_outage_probability}
\end{align}
where $B$ is the transmission bandwidth and $\bar{R}$ is the target transmission rate. 
Minimizing the outage probability necessitates designing control systems to operate at reduced communication overhead.

%% file: Sec3_Prelimiaries.tex
\subsection{Background Principles}

The Koopman operator provides a global linear representation of non-linear dynamics, enabling efficient state prediction for closed-loop control. 
The \gls{stl} translates natural language-based control rules into a formal mathematical form.
Brief overviews of both concepts are provided below.   

\subsubsection{Koopman Operator}
\label{Sec3_1}

Consider a real-valued measurement function  $\psi: \mathbb{R}^{P}  \rightarrow \mathbb{R}$, termed as an observable, residing within an infinite-dimensional Hilbert space. 
The Koopman operator $\mathcal{K}$, a linear operator in this space, evolves the observable such that~\cite{proctor2018generalizing} \vspace{-3pt}
\begin{align}
    \label{eq_Koopman_linear}
     \psi(\mathbf{x}_{i,k+1}) = \mathcal{K} \psi(\mathbf{x}_{i,k}).
\end{align}
%
%
The Koopman operator linearizes non-linear dynamics in an observable space but operates in an infinite-dimensional space, posing challenges.  
To address this, we identify an invariant subspace spanned by Koopman eigenfunctions ${\psi_1, \psi_2, \cdots, \psi_d}$, where $d$ is a finite positive integer, allowing for a finite-dimensional representation.
Applying the Koopman operator to this subspace keeps the system in this subspace.
Leveraging this linearity structure and linearly embedding control commands within observable space, we derive a global linear approximation of non-linear dynamics in~\eqref{eq_control_affine} as~\cite{mamakoukas2019local} 
\begin{align}
    \label{eq_Koopman_affine}
     \Psi( \mathbf{x}_{i,k+1} ) =  \mathbf{K}_{A} \,
     \Psi( \mathbf{x}_{i,k} ) + \mathbf{K}_{B}  \, \mathbf{u}_{i,k},
\end{align}
where $\Psi(\cdot) \in \mathbb{R}^{d}$ is a vector of Koopman eigenfunctions, $\mathbf{K}_{A} \in \mathbb{R}^{d \times d}$ is the state-based transition Koopman matrix, and $\mathbf{K}_{B} \in \mathbb{R}^{d \times q}$ is the control-based Koopman matrix. 
Although the Koopman operator allows global linear representations of non-linear dynamics within the latent space to utilize linear control and prediction techniques, discovering Koopman eigenfunctions from finite data remains challenging. 
To address this, we adopt a \gls{ae}-based deep learning approach among various deep learning architectures, such as \gls{ae} and transformers, due to its simplicity and computational efficiency in learning lower-dimensional representations of system states.

\subsubsection{Signal Temporal Logic}
\label{Sec3_2}
\gls{stl} is a formal mathematical language to specify the temporal properties of discrete-time signals in control systems~\cite{leung2020back}. 
It enables the expression of time-bounded safety and control properties, making it suitable to guide the behavior of control systems under dynamic, real-time constraints.
The \gls{stl} defines control rules over predicates such as $\mu_{c} := \mathbf{x}_{i}(p) < c$, where $c$ is a scalar. 
The syntax of a \gls{stl} formula is given by 
\begin{align}
    \label{eq_STL_syntax}
    \varphi  := & \mu_c \;  | \; \varphi \land \vartheta \; | \; \square_{\left[ a ,b \right]} \varphi \;,
\end{align}
where $a,b \in \mathbb{Z}_{+}$ are finite discrete-time bounds, with $0 \leq a < b $, and $\varphi$ and $\vartheta$ are \gls{stl} formulas. $|$ (pipe symbol) separates between \gls{stl} formulas, $\land$ (and) is the boolean operator, and $\square$ (always) is the temporal operator that specifies the temporal properties of signals within the time interval $\left[ a,b \right]$.

A timed trace signal $s_{i,k}$ records ordered sequence system state variables with timestamps.  
The satisfaction of an \gls{stl} formula $\varphi$ by a timed trace signal $s_{i,k}$ is quantified by the robustness function $\rho$, which measures how strongly the signal satisfies or violates the formula:~\cite{leung2020back} 
\begin{align}
 \label{eq_robustness_value}
    & \rho(s_{i,k}, \mu_{c})  \quad \quad = c - \mathbf{x}_{i}(p) \nonumber \\ 
    & \rho(s_{i,k}, \varphi \land \vartheta)  \; \; \;=  \min ( \rho(s_{i,k},\varphi),  \rho(s_{i,k},\vartheta) ) \nonumber \\ 
    & \rho(s_{i,k}, \square_{[a,b]} \varphi) \;  =  \min_{k' \in [k+a,k+b]}  \rho(s_{i,k'},\varphi). 
\end{align} 
The robustness function maps a timed trace and \gls{stl} formula to a real value: positive for satisfaction and negative for violation.
Using a parse tree for STL formulas, we generate a directed acyclic graph to compute the robustness efficiently, with each sub-graph representing a node in the \gls{stl} formula's operation sequence~\cite{leung2019backpropagation,girgis2023semantic}.
This resulting computation graph reflects the robustness of \gls{stl} formulas as described in~\eqref{eq_robustness_value}. 

%% file: Sec4_CLD_Koopman_AE.tex
\section{Semantic Logical Koopman Auto-encoder}
\label{Sec4}

\begin{figure*}[t]
    \centering
    \subfigure[First phase for training the encoder, decoder, and \gls{dsk} model.\label{fig_CLD1_1}]{\includegraphics[trim={1.3cm 5.5cm 1.1cm 1.2cm}, clip, width=0.44\textwidth]{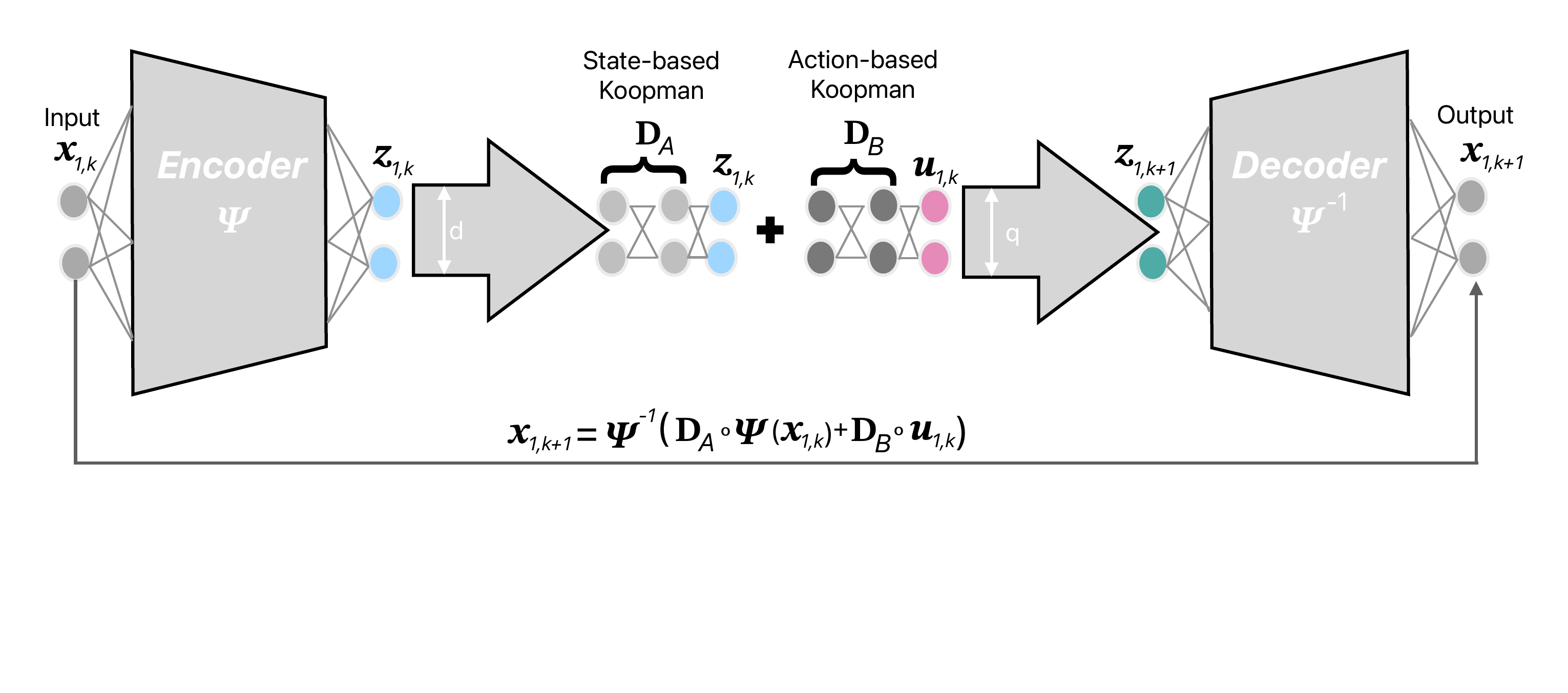} }
    \subfigure[Second phase for training the \gls{lsk} model.\label{fig_CLD1_2}]{\includegraphics[trim={1.3cm 5.5cm 1.1cm 1.2cm}, clip,width=0.44\textwidth]{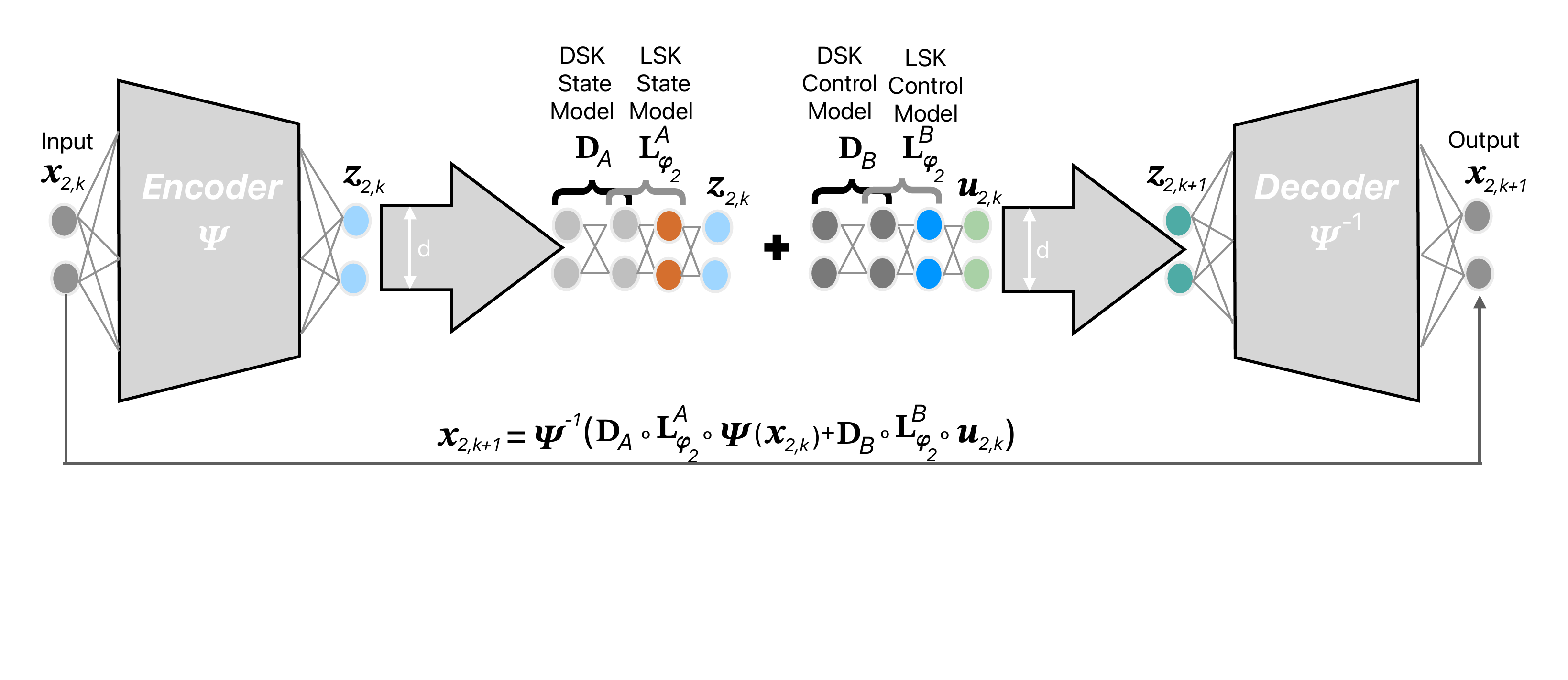}} \quad \quad
   \caption{ Training procedure of the semantic logical Koopman \gls{ae} for controlling two \gls{mld} systems with two main phases.}
    \label{fig_DSK_LSK}
\end{figure*} 


This section introduces a novel semantic \gls{cococo} for controlling multi-objective distinct control systems, utilizing a semantic logical Koopman \gls{ae} inspired by the adapter technique in~\cite{zhang2023llama}, which adapts a pre-trained model to new tasks without full retraining.
In this framework, the Koopman model incorporates a trainable \gls{lsk} model while keeping the \gls{dsk} model frozen. 
The architecture consists of an encoder, a decoder, and a Koopman model with state and action components. 
The Koopman model shown in Fig.~\ref{fig_DSK_LSK} is divided into four parts: two for the \gls{dsk} model (state and action matrices) and two for the \gls{lsk} model (state and action matrices), which govern the linear state evolution based on control rules.
The training procedure consists of two phases. 
In the first phase, the encoder, decoder, and \gls{dsk} model are jointly trained on a reference system, then, in the second phase, these components are transferred to other systems where the \gls{ae} is fine-tuned using a small subset of local data to adapt to system-specific dynamics. 
Concurrently, the \gls{lsk} model is trained to encode the local control rules via a temporal logic-based loss.  
Once trained, each system is monitored remotely through efficient linear predictions by the semantic logical Koopman \gls{ae}.

\subsection{Dynamic Semantic Koopman Model}
\label{sec4_1}

In the first phase of our remote control approach, a reference plant with simple dynamics and no specific control rule (denoted as $\varphi_i := \mathsf{NULL}$) is selected as a benchmark.
Here, we aim to optimize the encoder $\Psi(\cdot)$, decoder $\Psi^{-1}(\cdot)$, and \gls{dsk} model, represented by the matrices $\mathbf{D}_A$ and $\mathbf{D}_B$. 
The optimization problem is formulated as. 
\begin{align}  
\label{eq_main_Opt2_ab}
{\footnotesize
\underset{ \mathbf{D}_A, \mathbf{D}_B,  \Psi, \Psi^{-1} }{ \arg \min } \; \frac{1}{K_{s}} \sum_{k = 1}^{K_s} \left\lVert \mathbf{x}_{i,k+k^{\prime}} - \Psi^{-1} \left( \tilde{\mathbf{z}}_{i,k+k^{\prime}} \right) \right\rVert^{2}_{2},
}
\end{align}
with the future predicted state representation of the reference plant $i$ at time $k+k^{\prime}$ given as. 
\begin{align}
    \tilde{\mathbf{z}}_{i,k + k^{\prime}} = \mathbf{D}^{k^{\prime}}_A \mathbf{z}_{i,k} + \mathbf{D}^{k^{\prime}}_{B} \mathbf{u}_{i,k}, 
\end{align}
where $\mathbf{z}_{i,k} = \Psi(\mathbf{x}_{i,k})$ is the state representations of plant $i$ at time $k$.
Due to the complexity of obtaining an analytical solution of~\eqref{eq_main_Opt2_ab} for long-term prediction, we leverage deep learning and define three loss functions given as. 
\begin{enumerate}
     \item A \textit{reconstruction loss} ($\mathcal{L}_1$) ensures accurate reconstruction of the system states and is defined as. 
     \begin{align}
            \label{eq_Reconst_loss}
            \mathcal{L}_1 = \frac{1}{K_s} \sum_{k = 1}^{K_{s}} || \mathbf{x}_{i,k} - \Psi^{-1} \left(\mathbf{z}_{i,k} \right) ||^{2}_{2}, 
        \end{align} 
        where $K_{s}$ is the number of system states observed in the first phase of remote control. 

    \item A \textit{linear dynamics loss} ($\mathcal{L}_2$) enforces the linearity of system state evolution within the latent space, as follows. 
   \begin{align}
        \label{eq_Linear_Loss}
        \mathcal{L}_2 = \frac{1}{K_{s}}    \sum_{k = 1}^{K_s} ||  \mathbf{z}_{i,k+k'} - \tilde{\mathbf{z}}_{i,k+k^{\prime}}  ||^{2}_{2},
    \end{align} 
    where $\tilde{\mathbf{z}}_{i,k+k^{\prime}} = \Psi \left( \mathbf{x}_{i,k+k^{\prime}} \right)$ is the future state representation of plant $i$ at time $k+k^{\prime}$.

    \item A \emph{prediction loss} ($\mathcal{L}_3$) ensures accurate prediction of future system states and is provided as follows. 
    \begin{align}
        \label{eq_Pred_loss}
        \mathcal{L}_3 = \frac{1}{K_{s}} \sum_{k = 1}^{K_s} || \mathbf{x}_{k+k'} - \Psi^{-1}( \tilde{\mathbf{z}}_{i,k+k^{\prime}} ) ||^{2}_{2}.
    \end{align} 

\end{enumerate}

The total loss function combines these three losses with a $\mathit{l}_2$ regularization term to prevent overfitting, leading to the final optimization problem formulated as. 
\begin{align}
    \label{eq_Koopman_opt_problem}
    \underset{ \mathbf{D}_{A}, \mathbf{D}_{B}, \Psi,\Psi^{-1} }{\arg \min}  \sum_{n = 1}^{3} c_{n} \mathcal{L}_{n} + c_{4} \left\lVert \mathbf{W} \right\lVert^{2}_{2},
\end{align} 
where $c_n$ are positive real coefficients and $\mathbf{W}$ represents the model weights. 
During this phase, frequent wireless transmission of system states and their representations is required to solve the optimization problem in~\eqref{eq_Koopman_opt_problem}.

In the second phase, starting at time $t = k\tau_o$, after training the \gls{dsk} model, the sensor transmits state representations $\mathbf{z}_{i,k}$ to the remote controller, which computes control commands $\mathbf{u}_{i,k}$. 
The controller predicts future states at time $t = (k+l)\tau_o$ using the \gls{dsk} model and decoder as 
\begin{align}
    \label{eq_future_state_pred}
    \tilde{\mathbf{x}}_{k+l}  = \Psi^{-1}(\mathbf{D}^{l}_{A} \Psi(\mathbf{x}_{i,k}) + \mathbf{D}^{l}_{B} \mathbf{u}_{i,k}),
\end{align}
for $l \in \mathbb{Z}_{+}$. 
To counter prediction errors from communication noise and system uncertainties, frequent model fine-tuning is required, after which other plants adopt the trained models for training their own models, outlined below.

\subsection{Logic Semantic Koopman Model} 
\label{sec4_2}

In this phase, we assume that all plants, except plant $i$, follow predefined control rules. 
The plant $j$ (where $j \in \mathcal{I} \backslash i$) operates under the control rule $\varphi_j$ and exhibits different dynamics from the plant $i$. 
Using the encoder, decoder, and \gls{dsk} model from plant $i$, plant $j$ derives an \gls{lsk} model (represented by matrices $\mathbf{L}^{A}_{\varphi_j}$ and $\mathbf{L}^{B}_{\varphi_j}$), which captures the system dynamics under $\varphi_j$. 
The optimization problem for the \gls{lsk} model is given as follows. 
\begin{align}  \label{eq_main_Opt3_ab}
  \underset{ \mathbf{L}^{A}_{\varphi_{j}}, \mathbf{L}^{B}_{\varphi_{j}} }{ \arg \min} \frac{1}{K_{s}}   \sum_{k = 1}^{K_s}  \left\lVert \mathbf{x}_{i,k+k^{\prime}} - \Psi^{-1} \left(  \tilde{\mathbf{z}}_{i,k+1}  \right)  \right\rVert^{2}_{2}.
\end{align} 
with the future predicted state representation of the plant $j$ at time $k+1$ given as \vspace{-5pt}
\begin{align}
    \tilde{\mathbf{z}}_{i,k+1} = \mathbf{L}^{A}_{\varphi_j}\mathbf{D}_{A} \mathbf{z}_{i,k} +  \mathbf{L}^{B}_{\varphi_j}\mathbf{D}_{B} \mathbf{u}_{i,k},
\end{align}
where $\mathbf{L}^{A}_{\varphi_{j}} $ and  $\mathbf{L}^{B}_{\varphi_{j}} $ represent the transformation matrices parameterized by the control rules $\varphi_{j}$.
Due to the challenges in obtaining an analytical solution, deep learning is employed. 
A new loss function, \emph{logic loss} ($\mathcal{L}_4$), ensures that state representations follow the control rule $\varphi_j$ and is given by \vspace{-4pt}
\begin{align}
    \label{eq_stl_loss}
    \mathcal{L}_4 = \textrm{ReLu}(-\rho(s_{j,k},\varphi_j)),
\end{align}
where $s_{j,k}$ denotes the timed trace signal of first state representations, $\varphi_j$ is the control rule expressed as an STL formula in the latent space. 

To develop the \gls{stl} formula, we first train a conventional \gls{ae} for a few epochs, then fine-tune the \gls{lsk} model using time-series state representations and the predefined \gls{stl} template. 
For further details, we refer the reader to~\cite{girgis2023semantic}.
The final training objective combines the four loss functions ($\mathcal{L}_1, \mathcal{L}_2, \mathcal{L}_3, \mathcal{L}_4$) with $l_2$ regularization to prevent overfitting as follows. \vspace{-5pt} 
\begin{align}
\label{eq_LSM}
    \underset{\mathbf{L}^{A}_{\varphi_j}, \mathbf{L}^{B}_{\varphi_j} }{ \arg\min} \sum_{n = 1}^{4} c_{n}' \mathcal{L}_{n}  + c_{5}' \left\lVert \mathbf{W} \right\lVert^{2}_{2}.
\end{align}
Once the \gls{lsk} model is fully trained, plant $j$ integrates it with the \gls{dsk} model to construct its Koopman model and perform remote control operations.

%% file: Sec5_Simulation_Results.tex
\section{Simulation Results}
\label{Sec5}

This section validates the semantic logical Koopman \gls{ae} on inverted cart-pole systems over wireless channels.
Each system is defined by a four-dimensional state vector representing the horizontal position and velocity of the cart and the vertical angle and angular velocity of the pendulum, and control is applied through a horizontal force.
Here, we adopt the following simulation parameter settings unless specified otherwise: the pendulum mass is set to $\{1, 3, 4, 5, 6\} \, \mathrm{Kg}$, the cart mass is established at $ \{5, 15, 20, 25, 30\}\, \mathrm{Kg}$, and the pendulum length is maintained at $ \{ 0.2, 0.6, 0.8, 1.0, 1.2 \}\, \mathrm{m}$.
We conduct simulations on five distinct systems, each characterized by varying parameters that change their underlying dynamics, and governed by logical control rules, over a temporal interval $[150,251]$.
These logical control rules are defined as \vspace{-5pt}  
\begin{align}
    \label{eq_diff_environemnts}
      &\varphi_{i} := \square_{[151,251]}( ( s_{i,k} \geq \beta_{i} ) \,\land\, ( s_{i,k} \leq \beta_{i}) ), 
\end{align}
for $i \in \{1,2,3,4,5\}$. 
The parameter $\beta_{i}$, ranging across $\{0, 2.0, 3.0, 4.0, 5.0\}$, specifies the value within the state space for each respective system. 
The training data is sampled at a rate of $\tau_{0} = 100 \, \mathrm{ms}$ over a trajectory length ($K$) of $\left[0,251\right]$, and the system dynamics is computed using the Runge-Kutta method with a step size of $0.1$.
The semantic logical Koopman \gls{ae} is trained using the Adam optimizer with a batch size of $256$, a learning rate of $2 \times 10^{-3}$, and early stopping, and the experiments are repeated five times with different seeds.

The encoder consists of a fully connected layer with $h_{1} = 32$ neurons, activated by a ReLU function, followed by an output layer with $d \in \{2, 4\}$ neurons and sigmoid activation. 
The decoder mirrors the encoder, while the \gls{dsk} model includes two fully connected layers, and the \gls{lsk} model adds two more layers after the \gls{dsk} matrices are applied. 
The model is trained with weight hyperparameters $c_1 =  0.1$, $c_4 = 10^{-6}$, and $c_{2} = c_{3} = 0.75$ for~\eqref{eq_Koopman_opt_problem}, and $c'_1 = 0.1$, $c'_5 = 10^{-6}$, $c'_2 = c'_3 = 0.75$, and $c'_4 = 0.1$ for~\eqref{eq_LSM}.
Wireless performance is evaluated based on \gls{snr} values $\gamma_{i,k} \in \{5, 15 \} \, \mathrm{dB}$ with a transmission bandwidth ($B$) of  $20 \; \mathrm{MHz}$, a path loss exponent ($\eta$) of $3.0$, a transmission power ($P$) of $0.1 \, \mathrm{Watt}$, and  the distance between communication pairs ($D$) of $ 50 \, \mathrm{m}$.

The performance is evaluated by \textit{sample complexity} (number of communication samples required to train the model) and \textit{ prediction performance} using \gls{nrmse}, calculated as \vspace{-3pt}
\begin{align}
    \label{eq_NRMSE}
    \mathcal{N}_{i,K_p} = \frac{\sqrt{ \frac{1}{K_{p}} \sum_{k = K_s + 1}^{K_s + K_p } \| \tilde{\mathbf{x}}_{i,k} - \mathbf{x}_{i,k} \|^{2}_{2} }}{\| \max (\mathbf{x}) - \min (\mathbf{x}) \|_{2}} \times 100 \%,
\end{align} 
where $K_p$ is the prediction horizon at the test.
The \textit{control performance} is measured by a scoring function that assigns a score of $1$ when the cart’s position is within $0.2$ units of the target location and the pendulum's angle remains within $0.05$ radians of the upright position, and $0$ otherwise.
We compare the semantic logical Koopman \gls{ae} with two baselines: 
\begin{itemize}
    \item Individual MLD-Koopman model (Baseline 1), where each system has its own \gls{ae}. Note that the MLD-Koopman model is from our preliminary work \cite{girgis2023semantic}.
    \item Shared MLD-Koopman model (Baseline 2) used across all systems.
\end{itemize}

\begin{figure}[t]
    \centering
    \subfigure[ \label{fig_latent_dim_4}]{\includegraphics[width=0.24\textwidth]{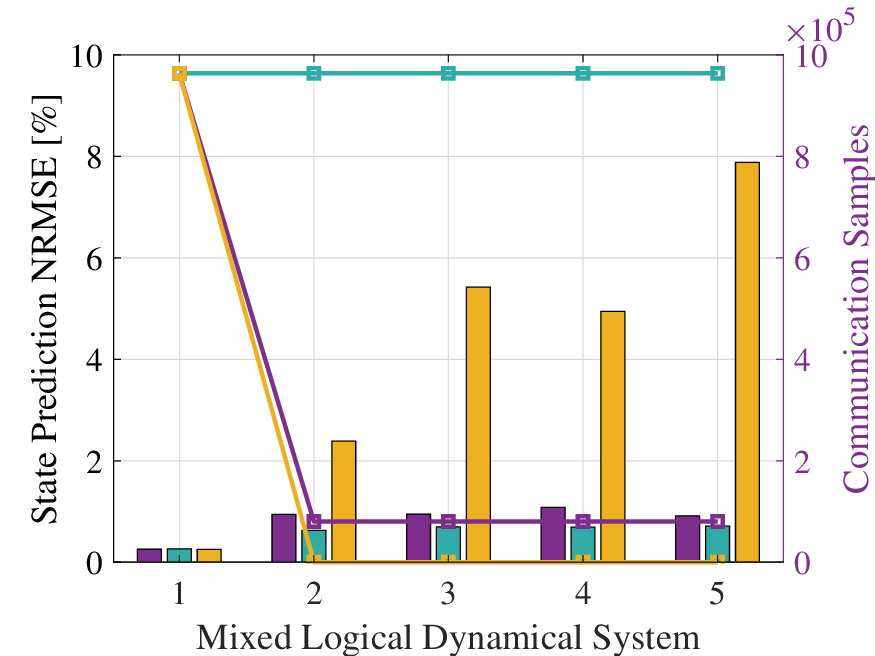}}
\subfigure[\label{fig_latent_dim_2}]{\includegraphics[width=0.24\textwidth]{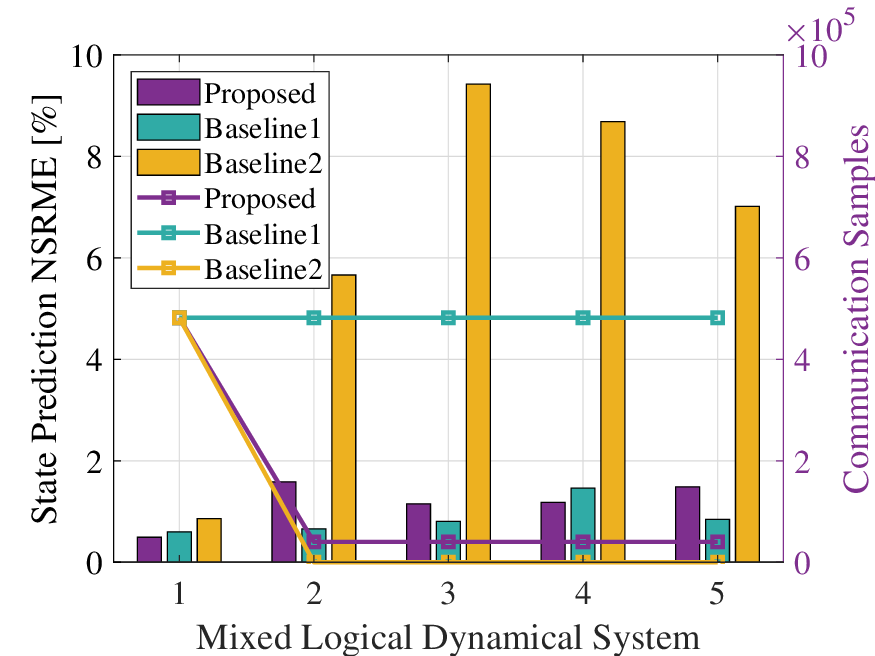}} 
   \caption{State prediction and communication samples for each system of proposed and baselines with $\text{SNR} =15$ dB for (a) $d = 4$ and (b) $d = 2$.}
    \label{fig_latent_dim}
\end{figure}

\textbf{Impact of Representation Dimensions.}\; Fig. \ref{fig_latent_dim} illustrates the state prediction performance and sample complexity of the proposed semantic logical Koopman \gls{ae} in comparison with two baselines, evaluated under an SNR of $15 \, \mathrm{dB}$, a one-step prediction depth, and varying state representation dimensions. 
The semantic logical Koopman \gls{ae} achieves performance comparable to Baseline 1 and significantly outperforms Baseline 2 across all systems. 
As shown in Fig.~\ref{fig_latent_dim_4}, the proposed model achieves an \gls{nrmse} of about $1\%$, with the reference system (system 1) reaching $0.2\%$ with a four-dimensional representation. 
Reducing the representation to two dimensions slightly increases the \gls{nrmse} to $1.5\%$ as shown in Fig.~\ref{fig_latent_dim_2}, while baseline 2 exhibits a substantially higher error of $9 \%$.

The improved performance of the proposed model is attributed to its ability to capture semantic structure through the \gls{dsk} model trained on the reference system, combined with the \gls{lsk} model, which encodes system-specific control rules in the latent space.  
The encoder and decoder are pre-trained for $50$ epochs using only $10\%$ of the data to train the \gls{lsk} model compared to the \gls{dsk} model. 
By jointly leveraging the \gls{dsk} and \gls{lsk} models, the proposed approach effectively linearizes non-linear system dynamics, enabling accurate state prediction with significantly reduced communication overhead.
Note that the encoder and decoder are fine-tuned on each target system to adapt to system-specific control rules, enriching the latent space to capture dynamic variability.

Although the proposed model reduces the communication cost by transmitting and predicting low-dimensional representations, it introduces a modest computational cost that stems from encoding logical control rules in the latent space.
Its training complexity combines two parts: the \gls{dsk} and \gls{lsk} models. 
Specifically, the \gls{dsk} training scales as $O(N.  \mathcal{E}. ( p. h_{1} + h_{1}.p + d^{2}) )$, where $\mathcal{E}$ is the number of epochs during \gls{dsk} training and $N$ is the number of training samples, while the \gls{lsk} training scales as $O(0.1  N. \mathcal{E}_{l}. d^{2})$, where $\mathcal{E}_l$ is the number of epochs during \gls{lsk} training. 
Hence, the overall training complexity is $O(N.  \mathcal{E}. ( p. h_{1} + h_{1}.p + d^{2}) + 0.1  N. \mathcal{E}_{l}. d^{2})$. 
During the encoding of logical control, the operational time complexity of encoding control rules over trajectories of length $K$ scales as $O(0.1 N.\mathcal{E}_{l}. K. |\varphi|)$, where $|\varphi|$ is the number of operators in the logical control formula.   
Despite this computational overhead, the proposed model is more efficient than Baseline $2$, which requires training a Koopman \gls{ae} separately for each system, incurring per-system cost comparable to the \gls{dsk} model training. 
In contrast, the proposed model trains the \gls{dsk} model once and, for each new system, incurs only the small additional cost of training the lightweight \gls{lsk} model. 
This trade-off enables efficient scalability across heterogeneous systems while maintaining low communication and computation requirements.   

As shown in Fig.~\ref{fig_latent_dim_4}, high-dimensional representations enable the proposed model to capture dominant Koopman eigenfunctions, improving state prediction performance. 
In contrast, reducing the representation dimension from four to two reduces the number of transmitted samples by approximately half while maintaining acceptable prediction accuracy.

\begin{figure}[t]
    \centering
    \subfigure[ \label{fig_score_4D}]{\includegraphics[width=0.24\textwidth]{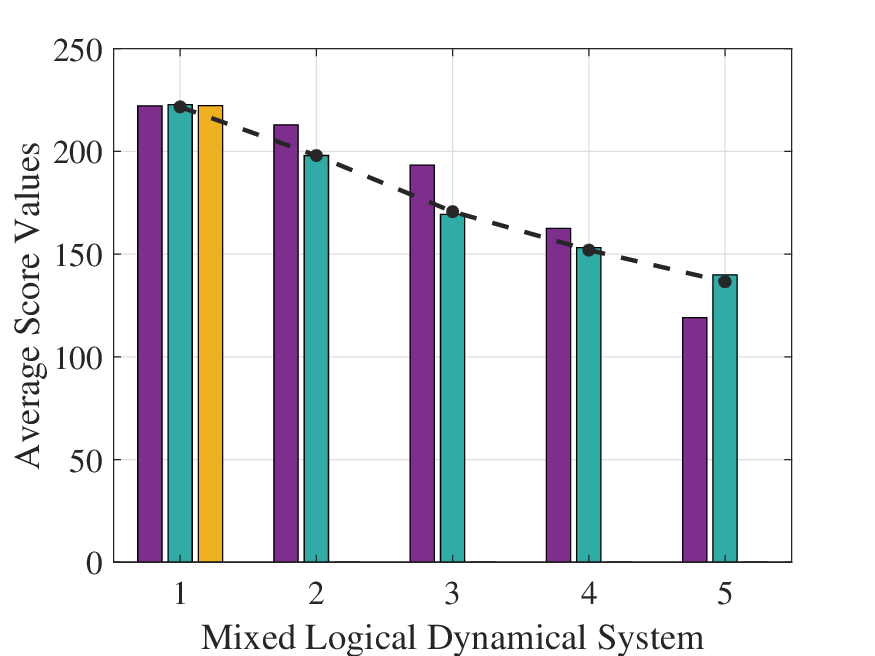}}
\subfigure[\label{fig_score_2D}]{\includegraphics[width=0.24\textwidth]{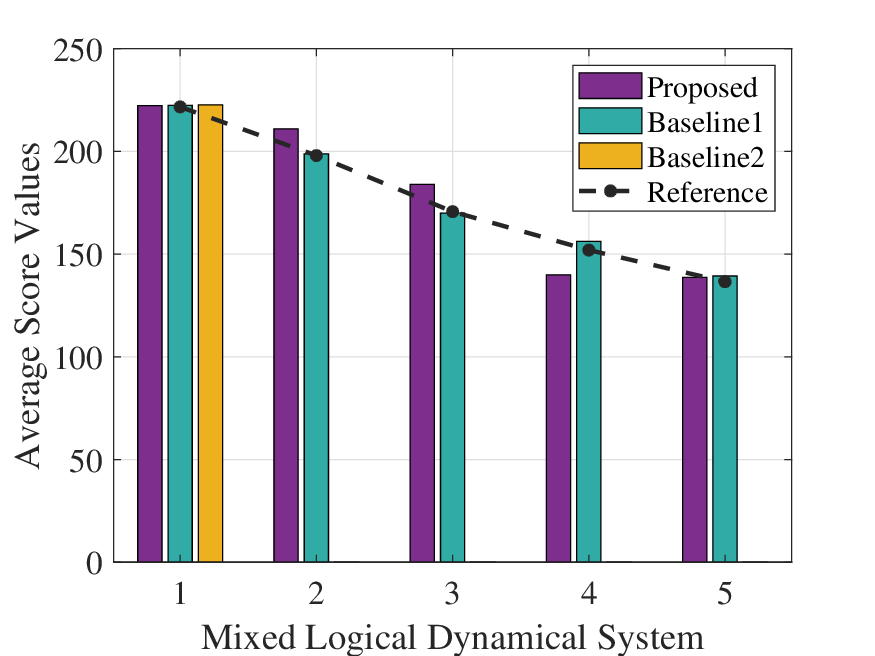}} 
   \caption{ Average score values for each system of the proposed and baselines with $\text{SNR} =15$ dB for (a) $d = 4$ and (b) $d = 2$.}
    \label{fig_score_dim}
\end{figure}

\textbf{Control Performance.}\; Fig. \ref{fig_score_dim} presents the control performance of the proposed model and two baselines, evaluated in terms of the average score values under varying representation dimensions and \gls{snr} of $15 \, \mathrm{dB}$. 
The proposed model achieves control performance closely aligned with the reference values across all systems, similar to baseline 1, which maintains high prediction accuracy with approximately $0.5\%$ NRMSE for four-dimensional and $1\%$ for two-dimensional representations.
In contrast, baseline 2 exhibits lower prediction accuracy, which negatively impacts its control performance.

Although reducing the representation dimensions to two leads to a slight decline in prediction accuracy, the control performance remains nearly unaffected, as shown in Fig.~\ref{fig_score_2D}. 
This result suggests that lower-dimensional representations can substantially reduce communication overhead without compromising control performance. 
Furthermore, average score values decrease for systems that deviate more significantly from the reference system, as these systems require more time to reach the desired state, leading to control performance degradation.   

\begin{figure}[t]
    \centering
    \subfigure[ \label{fig_15_dB}]{\includegraphics[width=0.24\textwidth]{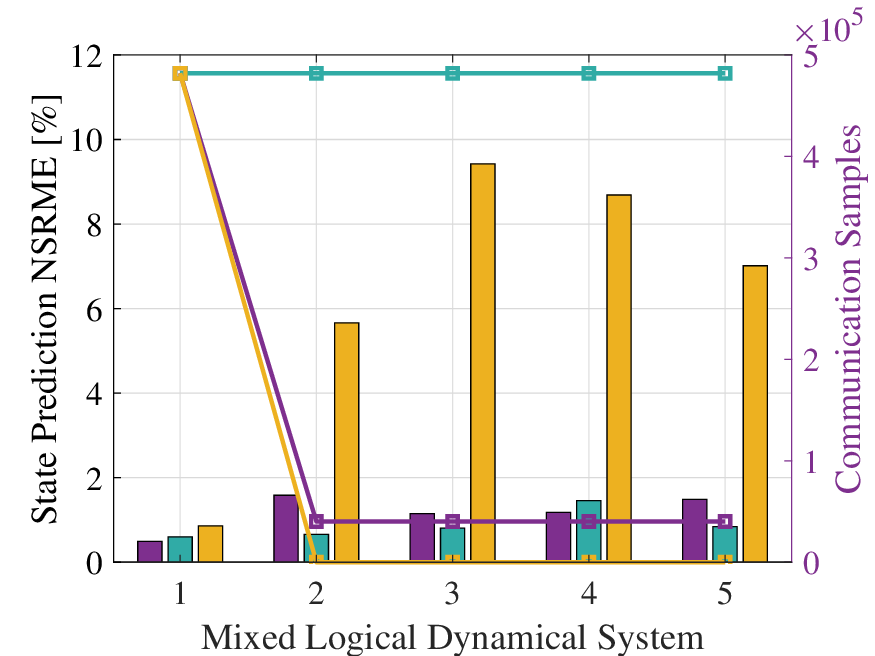}}
\subfigure[\label{fig_5_dB}]{\includegraphics[width=0.24\textwidth]{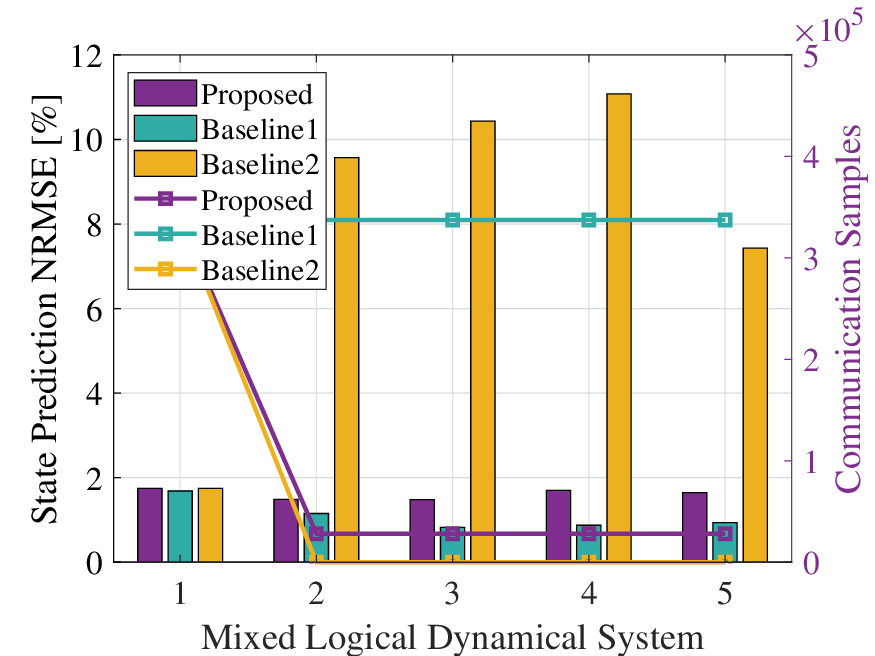}} 
   \caption{State prediction and communication samples for each system of the proposed and baselines for (a) $\text{SNR} =15$ dB and (b) $\text{SNR} =5$ dB.}
    \label{fig_SNR}
\end{figure}

\textbf{Impact of SNR.}\; Fig.~\ref{fig_SNR} illustrates the state prediction performance and sample complexity of the proposed model compared to two baselines for different \gls{snr} values, two-dimensional representations, and one-step prediction depth.
The results indicate that the prediction accuracy slightly decreases with lower \gls{snr} values for all models, mainly due to increased packet loss at $5 \, \mathrm{dB}$, which reduces the training data.
Interestingly, baseline 1 maintains comparable improved performance for systems $4$ and $5$ under low \gls{snr} values. 
This behavior is likely attributed to a smaller and more balanced dataset that mitigates overfitting. 
In contrast, the proposed model exhibits a more noticeable decline in prediction accuracy at low SNR values, which can be attributed to error propagation within the \gls{dsk} and \gls{lsk} models. 

%% file: Sec6_Conclusion.tex
\section{Conclusion}
\label{Sec6}

In this work, we proposed a novel semantic logical Koopman \gls{ae} designed to control multi-objective distinct systems under limited communication and computational resources.
We employed two key models: the \gls{dsk} model to capture the shared semantic structure in a linear form within the latent space, and the \gls{lsk} model to encode system-specific control rules within the latent space. 
Simulation results validate that the proposed approach's ability to scale across multi-objective distinct systems without compromising control performance. 

%% file: main.bbl
\begin{thebibliography}{10}
\providecommand{\url}[1]{#1}
\csname url@samestyle\endcsname
\providecommand{\newblock}{\relax}
\providecommand{\bibinfo}[2]{#2}
\providecommand{\BIBentrySTDinterwordspacing}{\spaceskip=0pt\relax}
\providecommand{\BIBentryALTinterwordstretchfactor}{4}
\providecommand{\BIBentryALTinterwordspacing}{\spaceskip=\fontdimen2\font plus
\BIBentryALTinterwordstretchfactor\fontdimen3\font minus
  \fontdimen4\font\relax}
\providecommand{\BIBforeignlanguage}[2]{{%
\expandafter\ifx\csname l@#1\endcsname\relax
\typeout{** WARNING: IEEEtran.bst: No hyphenation pattern has been}%
\typeout{** loaded for the language `#1'. Using the pattern for}%
\typeout{** the default language instead.}%
\else
\language=\csname l@#1\endcsname
\fi
#2}}
\providecommand{\BIBdecl}{\relax}
\BIBdecl

\bibitem{zeng2019joint}
T.~Zeng, O.~Semiari, W.~Saad, and M.~Bennis, ``Joint communication and control
  for wireless autonomous vehicular platoon systems,'' \emph{IEEE Transactions
  on Communications}, vol.~67, no.~11, pp. 7907--7922, 2019.

\bibitem{getu2024survey}
T.~M. Getu, G.~Kaddoum, and M.~Bennis, ``A survey on goal-oriented semantic
  communication: Techniques, challenges, and future directions,'' \emph{IEEE
  Access}, 2024.

\bibitem{girgis2021predictive}
A.~M. Girgis, J.~Park, M.~Bennis, and M.~Debbah, ``Predictive control and
  communication co-design via two-way {G}aussian process regression and
  {AoI}-aware scheduling,'' \emph{IEEE Transactions on Communications},
  vol.~69, no.~10, pp. 7077--7093, 2021.

\bibitem{eisen2022communication}
M.~Eisen, S.~Shukla, D.~Cavalcanti, and A.~S. Baxi, ``Communication-control
  co-design in wireless edge industrial systems,'' in \emph{2022 IEEE 18th
  International Conference on Factory Communication Systems (WFCS)}.\hskip 1em
  plus 0.5em minus 0.4em\relax IEEE, 2022, pp. 1--8.

\bibitem{zhao2018toward}
G.~Zhao, M.~A. Imran, Z.~Pang, Z.~Chen, and L.~Li, ``Toward real-time control
  in future wireless networks: Communication-control co-design,'' \emph{IEEE
  Communications Magazine}, vol.~57, no.~2, pp. 138--144, 2018.

\bibitem{seo2021semantics}
H.~Seo, J.~Park, M.~Bennis, and M.~Debbah, ``Semantics-native communication via
  contextual reasoning,'' \emph{IEEE Transactions on Cognitive Communications
  and Networking}, vol.~9, no.~3, pp. 604--617, 2023.

\bibitem{strinati2024goal}
E.~C. Strinati, P.~Di~Lorenzo, V.~Sciancalepore, A.~Aijaz, M.~Kountouris,
  D.~G{\"u}nd{\"u}z, P.~Popovski, M.~Sana, P.~A. Stavrou, B.~Soret
  \emph{et~al.}, ``Goal-oriented and semantic communication in 6{G} {AI}-native
  networks: The 6{G}-{GOALS} approach,'' in \emph{2024 Joint European
  Conference on Networks and Communications \& 6G Summit (EuCNC/6G
  Summit)}.\hskip 1em plus 0.5em minus 0.4em\relax IEEE, 2024, pp. 1--6.

\bibitem{xie2021deep}
H.~Xie, Z.~Qin, G.~Y. Li, and B.-H. Juang, ``Deep learning enabled semantic
  communication systems,'' \emph{IEEE Transactions on Signal Processing},
  vol.~69, pp. 2663--2675, 2021.

\bibitem{10849550}
Y.~Wang, H.~Han, Y.~Feng, J.~Zheng, and B.~Zhang, ``Semantic communication
  empowered 6{G} networks: Techniques, applications, and challenges,''
  \emph{IEEE Access}, vol.~13, pp. 28\,293--28\,314, 2025.

\bibitem{girgis2023semantic}
A.~M. Girgis, H.~Seo, J.~Park, and M.~Bennis, ``Semantic and logical
  communication-control co-design for correlated dynamical systems,''
  \emph{IEEE Internet of Things Journal}, 2023.

\bibitem{eisen2019control}
M.~Eisen, M.~M. Rashid, K.~Gatsis, D.~Cavalcanti, N.~Himayat, and A.~Ribeiro,
  ``Control aware communication design for time sensitive wireless systems,''
  in \emph{ICASSP 2019-2019 IEEE International Conference on Acoustics, Speech
  and Signal Processing (ICASSP)}.\hskip 1em plus 0.5em minus 0.4em\relax IEEE,
  2019, pp. 4584--4588.

\bibitem{lima2020learning}
V.~Lima, M.~Eisen, and A.~Ribeiro, ``Learning constrained resource allocation
  policies in wireless control systems,'' in \emph{2020 59th IEEE Conference on
  Decision and Control (CDC)}.\hskip 1em plus 0.5em minus 0.4em\relax IEEE,
  2020, pp. 2615--2621.

\bibitem{de2021age}
P.~M. de~Sant~Ana, N.~Marchenko, P.~Popovski, and B.~Soret, ``Age of loop for
  wireless networked control systems optimization,'' in \emph{2021 IEEE 32nd
  Annual International Symposium on Personal, Indoor and Mobile Radio
  Communications (PIMRC)}.\hskip 1em plus 0.5em minus 0.4em\relax IEEE, 2021,
  pp. 1--7.

\bibitem{liu2014multiobjective}
C.~Liu, X.~Xu, and D.~Hu, ``Multiobjective reinforcement learning: A
  comprehensive overview,'' \emph{IEEE Transactions on Systems, Man, and
  Cybernetics: Systems}, vol.~45, no.~3, pp. 385--398, 2014.

\bibitem{hayes2021practical}
C.~F. Hayes, R.~R{\u{a}}dulescu, E.~Bargiacchi, J.~K{\"a}llstr{\"o}m,
  M.~Macfarlane, M.~Reymond, T.~Verstraeten, L.~M. Zintgraf, R.~Dazeley,
  F.~Heintz \emph{et~al.}, ``A practical guide to multi-objective reinforcement
  learning and planning,'' \emph{arXiv preprint arXiv:2103.09568}, 2021.

\bibitem{brunton2016koopman}
S.~L. Brunton, B.~W. Brunton, J.~L. Proctor, and J.~N. Kutz, ``Koopman
  invariant subspaces and finite linear representations of nonlinear dynamical
  systems for control,'' \emph{PloS one}, vol.~11, no.~2, 2016.

\bibitem{leung2019backpropagation}
K.~Leung, N.~Ar{\'e}chiga, and M.~Pavone, ``Backpropagation for parametric
  stl,'' in \emph{2019 IEEE Intelligent Vehicles Symposium (IV)}.\hskip 1em
  plus 0.5em minus 0.4em\relax IEEE, 2019, pp. 185--192.

\bibitem{kaiser2021data}
E.~Kaiser, J.~N. Kutz, and S.~L. Brunton, ``Data-driven discovery of koopman
  eigenfunctions for control,'' \emph{Machine Learning: Science and
  Technology}, vol.~2, no.~3, p. 035023, 2021.

\bibitem{proctor2018generalizing}
J.~L. Proctor, S.~L. Brunton, and J.~N. Kutz, ``Generalizing {K}oopman theory
  to allow for inputs and control,'' \emph{SIAM Journal on Applied Dynamical
  Systems}, vol.~17, no.~1, pp. 909--930, 2018.

\bibitem{mamakoukas2019local}
G.~Mamakoukas, M.~Castano, X.~Tan, and T.~Murphey, ``Local koopman operators
  for data-driven control of robotic systems,'' in \emph{Robotics: science and
  systems}, 2019.

\bibitem{leung2020back}
K.~Leung, N.~Ar{\'e}chiga, and M.~Pavone, ``Back-propagation through signal
  temporal logic specifications: Infusing logical structure into gradient-based
  methods,'' in \emph{International Workshop on the Algorithmic Foundations of
  Robotics}.\hskip 1em plus 0.5em minus 0.4em\relax Springer, 2020, pp.
  432--449.

\bibitem{zhang2023llama}
R.~Zhang, J.~Han, A.~Zhou, X.~Hu, S.~Yan, P.~Lu, H.~Li, P.~Gao, and Y.~Qiao,
  ``Llama-adapter: Efficient fine-tuning of language models with zero-init
  attention,'' \emph{arXiv preprint arXiv:2303.16199}, 2023.

\end{thebibliography}
